\documentclass[11pt,a4paper]{article}
\usepackage{csquotes}
\usepackage[hyperref]{naaclhlt2018}
\usepackage{times}
\usepackage{microtype}
\usepackage{latexsym}
\usepackage{amsmath}
\usepackage{amsfonts}
\usepackage{verbatim}
\usepackage{soul}
\usepackage{url}
\usepackage{setspace}
\usepackage{graphicx}
\usepackage{multirow}
\usepackage{subcaption}
\usepackage{paralist}
\usepackage{gb4e}\noautomath

\aclfinalcopy 
\def\aclpaperid{420} 

\usepackage{booktabs}
\DeclareMathOperator*{\argmax}{argmax}
\usepackage{array}

\title{CliCR: A Dataset of Clinical Case Reports for Machine\\ Reading Comprehension\thanks{We provide the information about accessing the dataset, as well as the code for the experiments, at \url{http://github.com/clips/clicr}.}}

\author{Simon \v{S}uster \textnormal{and} Walter Daelemans\\
  Computational Linguistics \& Psycholinguistics Research Center,\\ University of Antwerp, Belgium \\
  {\tt \{simon.suster,walter.daelemans\}@uantwerpen.be}}
\date{}
\begin{document}
\maketitle
\begin{abstract}
We present a new dataset for machine comprehension in the medical domain. Our dataset uses clinical case reports with around 100,000 gap-filling queries about these cases. We apply several baselines and state-of-the-art neural readers to the dataset, and observe a considerable gap in performance (20\% F1) between the best human and machine readers. We analyze the skills required for successful answering and show how reader performance varies depending on the applicable skills. We find that inferences using domain knowledge and object tracking are the most frequently required skills, and that recognizing omitted information and spatio-temporal reasoning are the most difficult for the machines.

\end{abstract}

\section{Introduction}
Machine comprehension is a task in which a system reads a text passage and then answers questions about it. The progress in machine comprehension heavily depends on the introduction of new datasets \citep{Burges2013}, which encourages the development of new algorithms and deepens our understanding of the (linguistic) challenges that can or can not be tackled well by these algorithms.  Recently, a number of reading comprehension datasets have been proposed (\autoref{sec:related}), differing in various aspects such as mode of construction, answer-query formulation and required understanding skills. Most are open-domain datasets built from news, fiction and Wikipedia texts. For specialized domains, however, large machine comprehension datasets are extremely scarce \citep{WelblEtAl2017}, and the required comprehension skills poorly understood. With our work we hope to narrow this gap by proposing a new resource for reading comprehension in the clinical domain, and by analyzing the different types of comprehension skills that are triggered while answering \citep{SugawaraEtAl2017a,LaiEtAl2017}.

\begin{figure}
\small
\begin{tabular}{p{1.01\columnwidth}}
\textbf{passage:}\newline
[\ldots] A gradual improvement in clinical and laboratory status was achieved within 20 days of antituberculous treatment . The patient was then subjected to a thoracic CT scan that also showed significant radiological improvement . \textit{Thereafter , tapering of corticosteroids was initiated with no clinical relapse .} The patient was discharged after being treated for a total of 30 days and continued receiving antituberculous therapy with no reported problems for a total of 6 months under the supervision of his hometown physicians . [\ldots]\\
\textbf{query:}\newline
If steroids are used , great caution should be exercised on their gradual tapering to avoid \underline{\hspace{1cm}} .\\
\textbf{answer:}\newline 
relapse  (sem\_type=problem, cui=C0035020)\\
\end{tabular}
\caption{An example from the dataset, with the passage sentence relevant for answering italicized. The passage has been shortened for clarity.}\label{fig:ex}
\end{figure}

Machine comprehension for healthcare and medicine has received little attention so far, although it offers great potential for practical use.
A typical application would be clinical decision support, where given a massive amount of text, a clinician asks questions about either external, medical knowledge (reading literature) or about particular patients (reading electronic health records). Currently, patient-specific questions are tackled by manually browsing or searching those records. This task can be facilitated by summarization and QA systems \citep{DemnerFushmanAndLin2007,DemnerFushmanEtAl2009}, and we believe, by fine-grained machine reading. Reading comprehension systems that perform on a finer level could play an important role especially when combined with document retrieval to perform machine reading at scale, such as in the models of  \citet{ChenEtAl2017} and \citet{WatanabeEtAl2017} for the general domain.\looseness=-1

For our dataset, we construct queries, answers and supporting passages from BMJ Case Reports, the largest online repository of such documents. A case report is a detailed description of a clinical case that focuses on rare diseases, unusual presentation of common conditions and novel treatment methods. Each report contains a \textit{Learning points} section, summarizing the key pieces of information from that report. The learning points are typically paraphrased portions of passage text and do not match passage sentences exactly. We use these learning points to create queries by blanking out a medical entity. To counteract potential errors and inconsistencies due to automated dataset creation, we perform several checks to improve the quality of the dataset (\autoref{sec:design}). Our dataset contains around 100,000 queries on 12,000 case reports, has long support passages (around 1,500 tokens on average) and includes answers which are single- or multi-word medical entities. We show an example from the dataset in Figure \ref{fig:ex}. 

We examine the performance on the dataset in two ways. First, we report machine performance for several baselines and neural readers. To enable a more flexible answer evaluation, we expand the answers with their respective synonyms from a medical knowledge base, and additionally supplement the standard evaluation metrics with BLEU and embedding-based methods. We investigate different ways of representing medical entities in the text and how this affects the neural readers. We obtain the best results with a recurrent neural network (RNN) with gated attention \citep{DhingraEtAl2017a}, but a simple approach based on embedding similarity proves to be a strong baseline as well. Second, we look at how well humans perform on this task, by asking both a medical expert and a novice to answer a portion of the validation set. When categorizing the skills necessary to find the right answer, we observe that a large number of comprehension skills get activated and that prior knowledge in the form of the ability to perform lexico-grammatical inferences matters the most. This suggests that for our dataset and possibly for domain-specific datasets more generally, more background knowledge should be incorporated in machine comprehension models. The current gap between the best machine and the best human performance is nearly 20\% F1, which leaves ample space for further study of machine readers on our dataset. In brief, the contributions of our paper are:
\begin{compactitem}
\item We propose a large dataset for reading comprehension in the medical domain, using clinical case descriptions.\looseness=-1
\item We carry out an empirical analysis of \begin{inparaenum}[\itshape a\upshape)] \item system and human performance on reading comprehension, and \item comprehension skills that are required for answering the queries correctly and that allow us to position the dataset according to its difficulty on each of the skills.\end{inparaenum}
\end{compactitem}

\section{Related datasets}\label{sec:related}
\begin{table}[t]
\small
\begin{tabular}{p{2.7cm} p{1.8cm} l r}
Dataset & \parbox[t]{5cm}{Question origin} &  Domain & Size \\
\cmidrule(lr){1-4}
\parbox[t]{5cm}{CliCR\\(this work)} & Learning points & Medical & 105K \\
\parbox[t]{5cm}{Quasar-S\\\citep{DhingraEtAl2017b}} & Definitions & Software & 37K \\ 
\parbox[t]{5cm}{SciQ\\ \citep{WelblEtAl2017}} & Crowdsourced & Science & 14K \\
\parbox[t]{5cm}{MedHop\\ \citep{WelblEtAl2017b}} & KB & Drugs & 2.5K \\
\parbox[t]{5cm}{Biology\\\citep{BerantEtAl2014}} & \parbox[t]{5cm}{Domain\\ expert} & Biology & 585 \\
\parbox[t]{5cm}{Algebra\\\citep{KushmanEtAl2014}} & Crowdsourced & Algebra & 514 \\
\parbox[t]{5cm}{QA4MRE\\\citep{SutcliffeEtAl2013}} & Annotator & Various & 240 \\
\end{tabular}\caption{Survey  of closed-domain reading comprehension datasets. Size: number of questions. We did not include remotely related datasets which concern a different task (e.g.\ information retrieval) \citep{RobertsEtAl2015,VoorheesAndTice2000}.}\label{tab:closeddomain}
\end{table}

Numerous \textbf{general-domain datasets} have been recently created to allow machine comprehension using data-intensive methods. These datasets were collected from
Wikipedia \citep{HewlettEtAl2016,JoshiEtAl2017,RajpurkarEtAl2016}, web search queries \citep{NguyenEtAl2016},
news articles \citep{HermannEtAl2015,OnishiEtAl2016,TrischlerEtAl2016},
books \citep{BajgarEtAl2016,HillEtAl2016,PapernoEtAl2016} and English exams \citep{LaiEtAl2017}. 
In Table~\ref{tab:closeddomain}, we compare our dataset to several \textbf{domain-specific datasets}  for machine comprehension. In Quasar-S, the queries are constructed from definitions of software entity tags in a community QA website, while in our case the queries are more varied and explicitly relate to the supporting passages. SciQ is a dataset of science exam questions, in which question-answer pairs are used to retrieve the text passages. For each question, four candidate answers are available. In our dataset, the number of candidate answer is much higher as the candidate answers come from the relatively long passages. Other datasets mentioned in the table are smaller, so they could not be used as training sets for statistical NLP models.

\textbf{Cloze datasets} require the reader to fill in gaps by relying on accompanying text.
Representative datasets are Children's Book Test \citep{HillEtAl2016} and Book Test \citep{BajgarEtAl2016}, in which queries are created by removing a word or a named entity from the running text in a book; and \citet{HermannEtAl2015}, who similarly to us blank out entities in abstractive CNN and Daily Mail summaries, but who are only concerned with short proper nouns and short passages. Who-did-what \citep{OnishiEtAl2016} requires the reader to select the person name from a short candidate list that best answers the query about a news event. They do not use summaries for query formation but remove a named entity from the initial sentence in a news article, and then perform information retrieval to find independent passages relevant to the query. Another cloze dataset for language understanding is ROCStories \citep{MostafazadehEtAl2016}, but it is targeted more towards script knowledge evaluation, and only contains five-sentence stories. Another related task is predicting rare entities only, with a focus on improving a reading comprehension system with external knowledge sources \citep{LongEtAl2017}.

Another popular way of creating datasets for reading comprehension is \textbf{crowdsourcing} \citep{RajpurkarEtAl2016,RichardsonEtAl2013,NguyenEtAl2016,TrischlerEtAl2016}. These datasets exist primarily for the general domain; for specialized domains where background knowledge is crucial, crowdsourcing is intuitively less suitable \citep{WelblEtAl2017b}, although some positive precedent exists for example in crowdsourcing annotations of radiology reports \citep{CocosEtAl2015}. 
Compared to automated dataset construction, crowdsourcing is more likely to provide high-quality queries and answers. On the other hand, human question generation may also lead to less varied datasets as questions would tend to be of \textit{wh-} type; for cloze datasets, the questions may be more varied and might require readers to possess a different set of skills.\footnote{Support for this is given in \citet{SugawaraEtAl2017a}, who show that Who-did-what dataset, for example, requires on average a larger number of reading skills than SQuAD \citep{RajpurkarEtAl2016} and MCTest \cite{RichardsonEtAl2013}.} 

\section{Dataset design}\label{sec:design}
We collected the articles from BMJ Case Reports\footnote{\url{http://casereports.bmj.com/}}. The data span the years 2005--2016 and amount to almost 12 thousand reports. We removed the HTML boilerplate from the crawled reports using jusText\footnote{\url{https://pypi.python.org/pypi/jusText}}, segmented and tokenized the texts with cTakes \citep{SavovaEtAl2010}, and annotated the medical entities using Clamp \citep{SoysalEtAl2017}. We apply two simple heuristics to refine the recognized entities and to decrease their sparsity. Namely, we move the function words (determiners and pronouns) from the beginning of the entity outside of it, and we adjust the entity boundary so that it does not include a parenthetical at the end of the entity. Clamp assigns entities following the i2b2-2010 shared task specifications \citep{UzunerEtAl2011}. For each entity, a concept unique identifier (CUI) is also available, which links it to the UMLS\textsuperscript{\textregistered} Metathesaurus\textsuperscript{\textregistered} \citep{LindbergEtAl1993}. To check the quality of the recognized entities, we carried out a small manual analysis on 250 entities. We found that in 89\% of cases, the boundaries were correct and defined a true entity. Wrongly recognized cases occurred mostly when two entities were coordinated and recognized as one; when a verb was wrongly included in the entity; or when a pre-modifier was left out.

\subsection{Query construction} We create a query by replacing a medical entity in one learning point with a blank. For example, in a report describing comorbid disorders of ADHD, we could obtain the following query:
\begin{exe}
\ex \enquote{Patients with ADHD have higher incidence of \underline{\hspace{0.6cm}}.}
\end{exe}
\noindent The missing entity \enquote{enuresis} is taken as the correct answer. Even though one query corresponds to at most one learning point, there can be more than one query built from a learning point. Occasionally, a learning point contains an exact repetition from the passage. These instances would be trivial to answer, so we remove them. We count as an exact match every instance whose longer side to left/right of the query blank coincides with a part in the passage text. This curation step reduces the dataset size by 5\%.
More commonly, the learning points are paraphrases of crucial parts of the passage. Sometimes, the entity answering the query is expressed differently in the passage. For example, in place of \enquote{enuresis}, the passage might include its synonym \enquote{bedwetting}. We manage these cases in two ways, by extending the set of answers for a certain query (\autoref{sec:answerset}), and adding a semantic relatedness metric to the standard evaluation (\autoref{sec:eval}).

\subsection{Answer set}\label{sec:answerset} We account for lexical variation of the ground-truth answers (compared to mentions in the passages) by extending each original ground-truth answer $a$ to a set of ground-truth answers $A$ using a knowledge base. Since our entity recognizer already provides the CUI labels, we can use them to obtain the list of alternative word and phrase forms (synonyms, abbreviations and acronyms) from UMLS\textsuperscript{\textregistered}.

Similarly to previous work \citep{ChoiEtAl2016,HewlettEtAl2016}, for certain queries none of the answers in $A$ occurs verbatim in the passage. We have found upon manual inspection that this is mostly due to lexical variation that is not captured by answer extension, and to a lesser degree, due to the introduction of entirely new information in the learning point and the entity recognition errors. In the empirical part, we use for training only the instances for which at least one answer occurs in the passage, but we evaluate on all instances in the validation and test sets, including those for which $A\cap E=\emptyset$, where $E$ is the set of all entities in the passage. This mimics a likely real-life scenario where the set of ground-truth answers is a priori unknown.\looseness=-1

\subsection{Task formulation}\label{sec:task}
The reading comprehension problem in our case can be represented as a tuple $(q,p,A)$, where $q$ is the query, built from a learning point; the passage $p$ is the entire report excluding the Learning points section; and $A$ is the set of ground-truth entities answering $q$. In defining the task, it is important to consider how to take into account entity annotation and how to define the answer output space. We look at these more closely in the rest of this section.
\begin{table}[t]
\centering
\resizebox{\columnwidth}{!}{%
\begin{tabular}{l r}
N of cases& 11,846\\
N of queries in train/dev/test& 91,344/6,391/7,184\\
N of tokens in passages & 16,544,217\\
N of word types in passages & 112,673\\
N of entity types in passages & 591,960 \\
N of distinct answers & 56,093\\
N of distinct answers (incl.\ extended) & 288,211\\
\% answers verbatim in passage & 59 \\
\end{tabular}%
}
\caption{Data statistics based on the lowercased dataset. For \textit{N of tokens in passages}, we count each passage exactly once, although several queries are normally associated with a passage.}\label{tab:stats}
\end{table}

Whenever the entities are marked in the passage, the system can learn to exploit this cue to find the answers more easily \citep{WangEtAl2017}. Although this simplifies the task, it also makes it less realistic as the entities may not be recognized at test time. Realizing that the presence of entities makes the task easier for the machines, \citet{HermannEtAl2015} anonymize the entities, also with a goal of discouraging language model solutions to the queries. In our case, it is not clear how relevant the anonymization is since we deal with medical entities, which have different properties than proper name entities \citep{KimEtAl2003,NiuEtAl2003}. We explore different entity-annotation choices in the empirical part, where we refer to them as \textbf{Ent} (entities marked) and \textbf{Anonym} (entities marked but anonymized). We further examine a more challenging setup in which the reader can not rely on entity markers as they are not present in the passage (\textbf{NoEnt}). In all cases, the reader chooses an answer among the candidates $E$ collected from all entities in the passage.\footnote{The candidate answers could in principle be obtained also in some other way, so we do not list them in our dataset.} 
Multi-word entities, which are common in our dataset, are treated as a single token by Ent and Anonym.\looseness=-1

\begin{figure}
\centering
\begin{subfigure}[b]{\columnwidth}
   \includegraphics[width=\columnwidth]{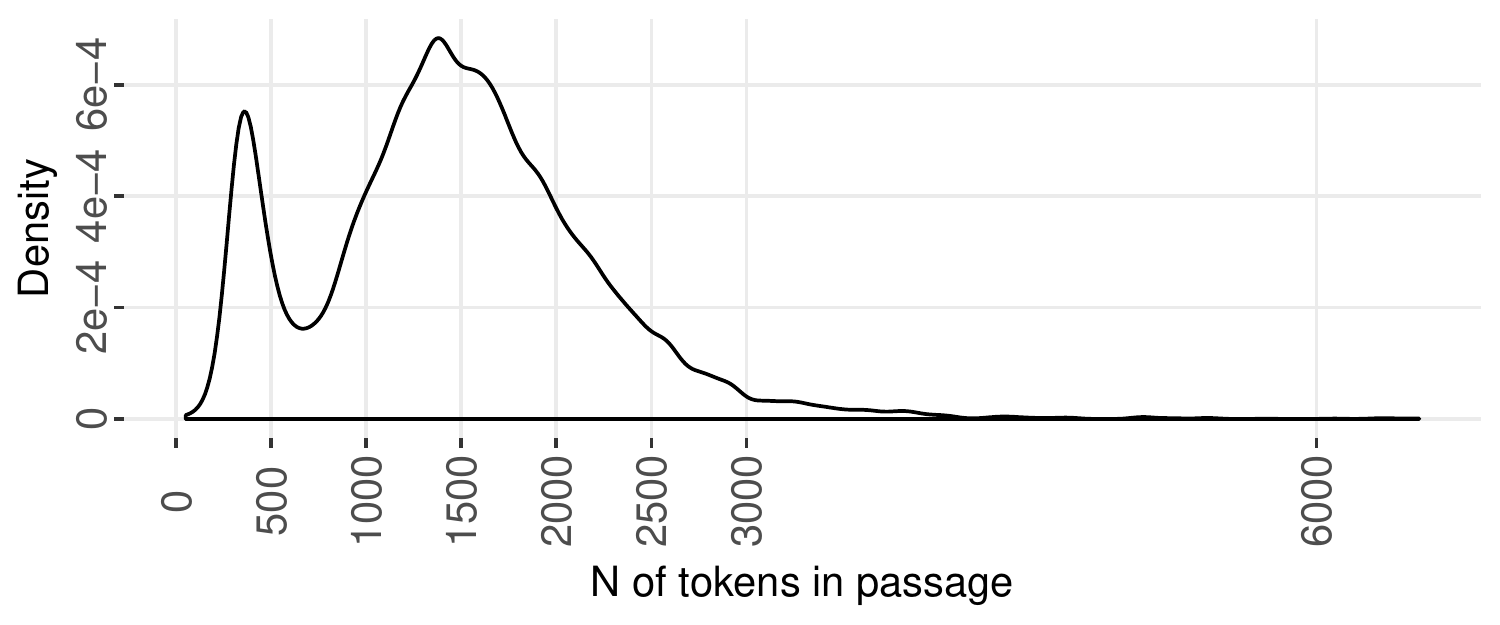}
   \caption{}
   \label{fig:plotlengthpassage} 
\end{subfigure}
\begin{subfigure}[b]{\columnwidth}
   \includegraphics[width=\columnwidth]{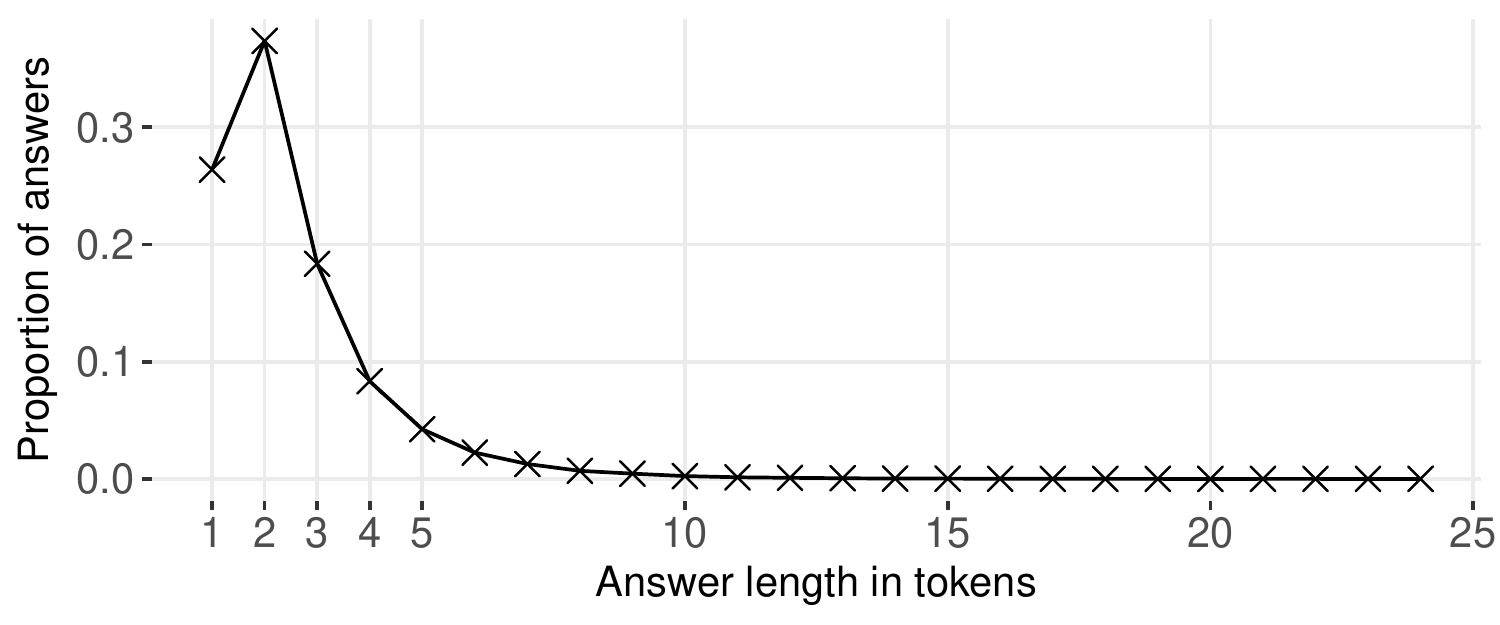}
   \caption{}
   \label{fig:plotlengthanswer}
\end{subfigure}
\caption{Distribution of (a) passage and (b) answer length. Curve (a) is bimodal due to shorter lengths of articles published prior to 2008.}
\end{figure}

\section{Dataset analysis}
\begin{figure}[t]
\begin{center}
\includegraphics[width=\columnwidth]{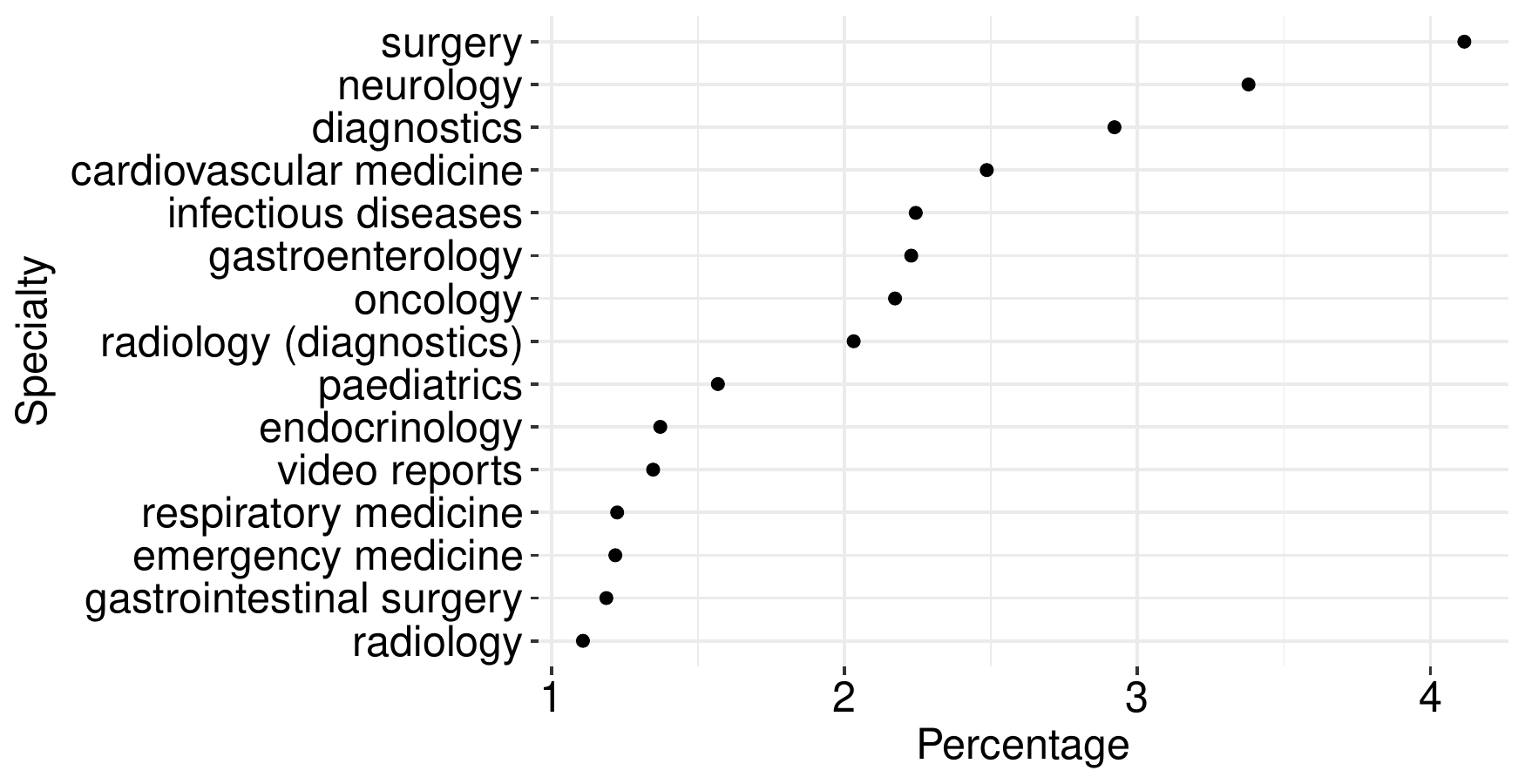}
\caption{The 15 most common medical specialties represented in the dataset.}
\label{fig:plotspecs}
\end{center}
\end{figure}

We now describe the dataset in more detail, starting with the general statistics summarized in Table~\ref{tab:stats}.
 It is worth pointing out that the support passages are rather long, which stems from the data origin (journal articles). We show the passage length distribution in Figure~\ref{fig:plotlengthpassage}, which has the average length of 1,466 tokens. Furthermore, passages are rich with medical entities. There is little repetition of answers---the total of around 100,000 queries are answered by 50,000 distinct entities. Upon extending the answer set with UMLS\textsuperscript{\textregistered} we introduce on average four alternative answers for each original one. In 59\% of instances, the answer entity is found verbatim in the relevant passage.  The answers can belong to any of the problem, treatment or test categories (Table \ref{tab:answertype}), and usually consist of multiple words (Figure \ref{fig:plotlengthanswer}). The diversity of medical specialties represented in the articles is shown in Figure~\ref{fig:plotspecs}.

\begin{table}[t]
\centering
\begin{tabular}{l r p{4.5cm}}
Type & \% & Example \\
\cmidrule(lr){1-3}
problem & 67 & tuberculosis, abdominal pain, acute myocardial infarction \\
treatment & 22	 & chemotherapy, surgical intervention, vitamin D suppl.\\
test & 11	& MRI, histopathological exam.\\
\end{tabular}\caption{Answer type statistics.}\label{tab:answertype}
\end{table}

\subsection{Analysis of comprehension skills}\label{sec:skills}
\begin{figure}[t]
\begin{center}
\includegraphics[width=\columnwidth]{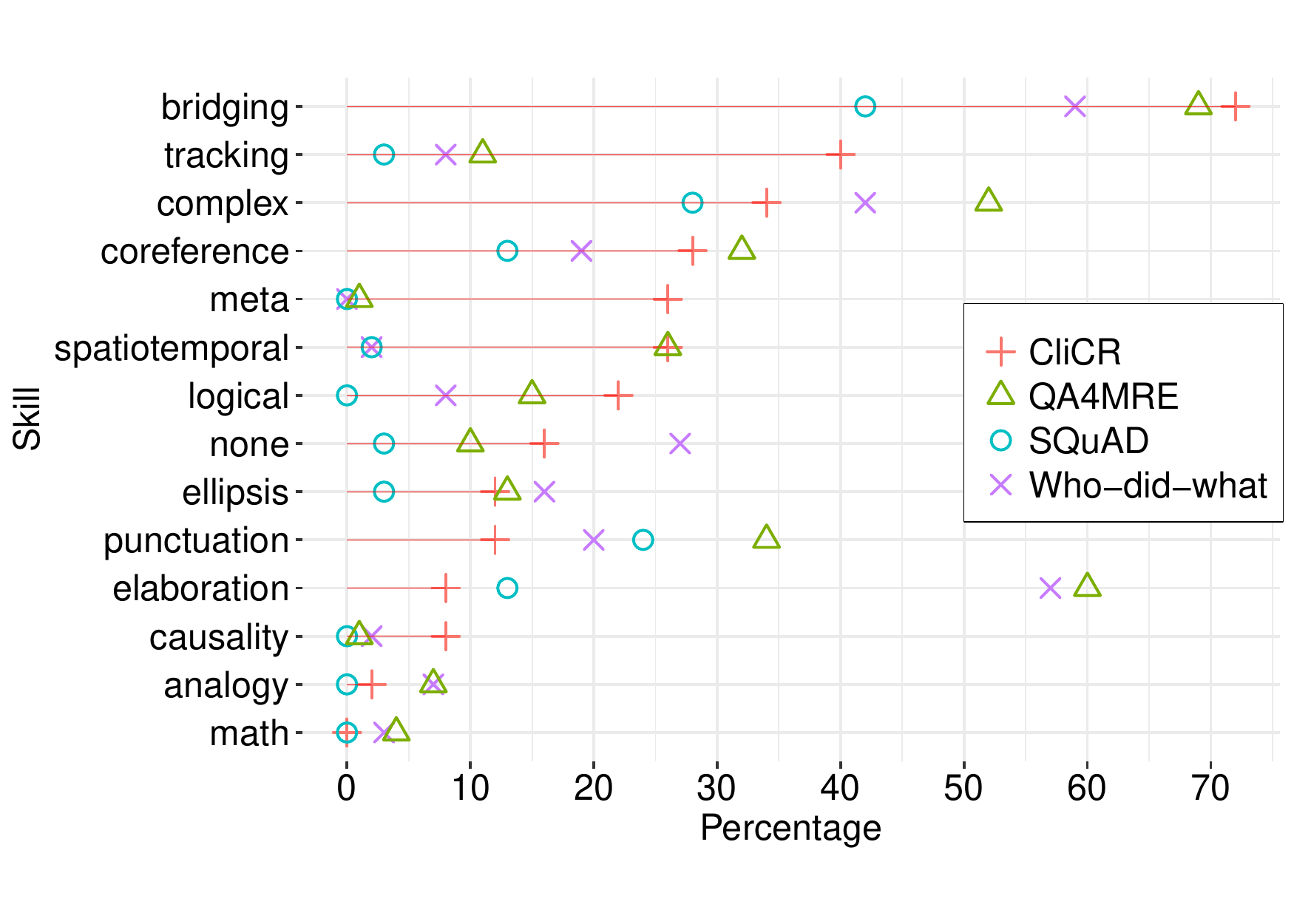}
\caption{Percentage of times a skill is required in a given dataset. The percentages for the datasets other than ours are from \citet{SugawaraEtAl2017a}.}
\label{fig:plotskills}
\end{center}
\end{figure}

We estimate the types of skills required in answering by following the categorization of \citet{SugawaraEtAl2017a}. We include the skill  definitions with examples from our dataset in \autoref{app:skills}. We annotated 100 instances in the validation set (with ground-truth answers provided), which yielded on average 2.85 skills per query. The distribution of the required skills is shown in Figure~\ref{fig:plotskills}. 
In comparison to the general-domain datasets (SQuAD, Who-did-what), our dataset and QA4MRE (which is also a domain-specific dataset, but with human-generated questions) require more bridging inferences (inferences using background knowledge about the domain), spatio-temporal reasoning and coreference resolution. In our dataset, meta knowledge and object tracking are required more often than in any other dataset. This can be explained by the data origin and the nature of queries. In the case reports, a prominent topic can be discussed which the author refers to in the query, but the query itself is never answered in the passage (meta knowledge). Furthermore, the authors often enumerate medical entities in the query, which leads to the frequent use of object tracking. The queries which were unanswerable are marked as \enquote{none}. The fraction of these cases was around 16\%. 

In our experience, the annotation of skills proved quite challenging due to certain confusables. For example, object tracking and coreference both need to maintain the link between objects; object tracking, which includes establishing set relations and membership, may be overlaid with the schematic clause relation skill (subordination); and bridging inference can overlap with coreference resolution. Nevertheless, we adhered to this classification of skills to increase comparability  to other datasets included in Figure~\ref{fig:plotskills}. 

\section{Methods}
\subsection{Baselines}
Our simplest baselines that we apply on the test set include choosing a random entity (\textbf{rand-entity}) and selecting the most frequent passage entity (\textbf{maxfreq-entity}) as the answer. We also include a distance-based method that uses word embeddings (\textbf{sim-entity)}. Here, we vectorize the passage and the query, and then choose that entity from the passage whose representation has the highest cosine similarity to the query representation:

\begin{equation}\label{eq:simentity}
\textit{sim-entity}=\argmax_{i\in E} \text{cos}\big(\sum_{j\in C_i} c_j,\sum_{k\in Q} q_k\big),
\end{equation}
\noindent where $c,q\in\mathbb{R}^d$.
The multiset $C_i$ contains the words $\{x_{i-n}, \ldots, x_{i-1}, x_{i+1}, \ldots, x_{i+n}\}$ surrounding the passage entity $i\in E$. We define $Q$, the context words of the query, likewise. 
To find out how well the queries can be answered without reading the passage, we also predict the most likely continuation with a language model (\textbf{lang-model}). We trained a 4-gram Kneser-Ney model on CliCR training data (with multi-word entities represented as a single token) using SRILM~\citep{Stolcke2002}.

\subsection{Neural readers}
We apply two types of bidirectional RNNs to our data. Following \citet{WangEtAl2017}, we distinguish between \textit{aggregation} readers and \textit{explicit reference} readers, which differ in their formulation of the attention mechanism and how it is being used for answer prediction. 

\paragraph{Stanford Attentive (SA) Reader} The model proposed by \citet{ChenEtAl2016} is an aggregation reader based on the Attentive Reader \citep{HermannEtAl2015}.  It predicts the answer using:
\begin{equation}\label{eq:sareaderpred}
\hat{a} = \argmax_{i\in E} e_o(i)^T o,
\end{equation}
where $e_o(i)$ is the answer's output embedding and $o$ is the passage representation obtained by weighting every token representation in the passage with attention: $o=\sum_t \alpha_t h_t$. The attention mechanism is used here to measure the compatibility between token ($h_t$) and query ($q$) representations with a bilinear form, $\alpha_t = \text{softmax}_t h_t^TW_\alpha q$. At prediction time, attention should highlight that position $t$ in the passage where the answer occurs. Note that the prediction relies on the aggregate representation $o$, hence the name  of the reader category. As we see in \eqref{eq:sareaderpred}, the prediction score does not allow accounting for multi-word entities, unless they are treated as a single token. Returning to our different set-ups based on entity annotation (\autoref{sec:task}), this means that we can apply SA reader with Ent and Anonym set-ups, but not with NoEnt, where multi-word answers should be allowed.

\paragraph{Gated-Attention (GA) Reader} \citet{DhingraEtAl2017a} investigate neural readers with a fine-grained attention mechanism that learns token representations for the passage that are also conditional on the query, but are in addition refined through multiple hops of the network. 
The model predicts the answer using attention weights with \textit{explicit reference} to answer positions in the passage:
\begin{equation}\label{eq:gareaderpred}
\hat{a} = \argmax_{i\in E} \sum_{t\in R(i,p)} \alpha_t,
\end{equation} where R is the set of indices in passage $p$ at which a token from the candidate $i$ occurs. This operation is also called the pointer sum attention \citep{KadlecEtAl2016}. Since the model marks the references for each token in the answer separately, it allows us to investigate also the NoEnt set-up.\footnote{We assume the candidate entities are known in advance.} 

We train each reader with the best hyper-parameters found on the validation set using random search \citep{BergstraAndBengio2012}, and evaluate it on the test part of the dataset. We provide more details about parameter optimization in \autoref{app:hyper}. The models use word embeddings pre-trained on biomedical texts. 

\subsection{Embedding data and pre-training}\label{optimization}
We induce the word embeddings on a combination of the CliCR training corpus and PubMed abstracts with open-access PMC articles available until 2015 (segmented and tokenized), amounting to over 9 billion tokens \citep{HakalaEtAl2016}. 
Considering the large effect of hyper-parameter selection on the quality of word embeddings \citep{LevyEtAl2015}, we optimize the embedding hyper-parameters also using random search.

\section{Evaluation}\label{sec:eval}

A model $f$ takes as input a passage--query pair and outputs an answer $\hat{a}$.\footnote{In our case, the answer is a word or a word phrase representing a medical entity. Alternatively, one could also take the UMLS\textsuperscript{\textregistered} CUI identifier as the answering unit. However, in that case, it would mean that sometimes the original word phrase is lost. This is because entity linking with CUIs can be noisy, and only a part of a word phrase may be linked to the ontology. In the current setup, we are able to keep both the original word phrase as well as the extended answers. The CUI information is still an integral part of the answer field in our dataset, so it can be used by other researchers if preferred.} We carry out the evaluation with different metrics described below. The final score $m$ for a metric $v$ is obtained by averaging over the test set:
\begin{equation}
m_{v}(f) = \frac{1}{|D_{\text{test}}|} \sum_{(p,q,A)\in D_{\text{test}}} \max_{a\in A} v(f(p,q), a).
\end{equation}
Since there are multiple correct answers $A$, we take the highest scoring answer $\hat{a}$ at each instance, as done in \citet{RajpurkarEtAl2016}. Note that in the dataset we do not supply the candidate answers; in the experiments, we constrain the candidates to the set of entities in the passage.

The two standardly used metrics for machine comprehension evaluation are the \textbf{exact match} (EM) and the \textbf{F1} score. For EM, the predicted and the ground truth answers must match precisely, safe for articles, punctuation and case distinction (same for other metrics). F1 metric is applied per instance and measures the overlap between the prediction $\hat{a}$ and the ground truth $a$, which are treated as bags of words.\footnote{In precision, the number of correct words is divided by the number of all predicted words. In recall, the former is divided by the number of words in the ground-truth answer.} 
While these two metrics are arguably sufficient in news-style machine comprehension where the entities are proper nouns which allow for little variation and synonymy, in our case the medical entities are often mostly common nouns modified by specifiers and qualifiers. To take into account potentially large lexical and word-order variation, we use two additional metrics.
First, we measure \textbf{BLEU} \citep{PapineniEtAl2002} for n-grams of length 2 (shortly, B-2) and 4 (B-4) using the package by \citet{ChenEtAl2015}, with which we aim to capture contiguity of tokens in longer answers. Second, it may occur that answers contain no word overlap yet still be good candidates because of their semantical relatedness, as in  \enquote{renal failure}--\enquote{kidney breakdown}. We take this into account by using an \textbf{embedding metric} (E-avg), in which we construct mean vectors for both ground-truth and system answer sequences, and then compare them with the cosine similarity. This and other embedding metrics for evaluation were previously studied in dialog-system research \citep{LiuEtAl2016}.

\section{Results and analysis}
\begin{table}[t]
\centering
\begin{tabular}{l c c c c c}
Method & EM & F1 & B-2 & B-4 & E-avg \\ 
\cmidrule(lr){1-6}
rand-entity & 1.4 & 5.1 & .03 & .01 & .23 \\ 
maxfreq-entity & 8.5 & 12.6 & .10 & .05 & .31 \\ 
sim-entity & 20.8 & 29.4 & .22 & .15 & .45 \\ 
lang-model & 2.1 & 3.5 & .00 & .00 & .30 \\
\cmidrule(lr){1-6}
SA-Anonym & 19.6 & 27.2 & .22 & .16 & .43 \\ 
SA-Ent & 6.1 & 11.4 & .07 & .05 & .31 \\ 
\cmidrule(lr){1-6}
GA-Anonym & 24.5 & 33.2 & .28 & .20 & .48  \\ 
GA-Ent & 22.2& 30.2 & .25 & .18 & .46  \\ 
GA-NoEnt & 14.9 & 33.9 & .21 & .11 & .51 \\
\cmidrule(lr){1-6}
\textit{human-expert}  & \textit{35} & \textit{53.7} & \textit{.46} & \textit{.23} & \textit{.67} \\ 
\textit{human-novice} & \textit{31} & \textit{45.1} & \textit{.43} & \textit{.24} & \textit{.62} \\ 
\end{tabular}%

\caption{Answering results on the test set. EM and F1 scores are percentages. The human scores (in italics) are based on the validation set.}\label{tab:test}
\end{table}

We show the results in Table~\ref{tab:test}. We see that answer prediction based on contextual representation of queries and passages (sim-entity) achieves a strong base performance that is only outperformed by GA reader. The language model performs poorly on EM and F1, but the embedding-metric score is higher, likely reflecting the fact that the predicted answers---though mostly incorrect---are related to the ground-truth answers. The poor performance means that based on queries alone (without reading the passage), it is difficult to provide accurate answers. 
The GA reader performs well across all entity set-ups, even when the entities are not marked in the passage. Interestingly, the exact match and BLEU scores in this case are much lower compared to other entity set-ups. Upon inspecting the predicted answers more closely, we have observed that GA-NoEnt tends to predict longer answers than GA-Ent/Anonym. For example, the average predicted answer length for GA-NoEnt was as high as 3.7 tokens, whereas for the other two set-ups and the ground-truth answers the numbers range between 2.3 and 2.5. A plausible explanation for this lies in how GA reaches its prediction \eqref{eq:gareaderpred}, which is by accumulating the attention weights without normalizing. This would then drive the model to prefer longer answers. For example, for the ground-truth entity \enquote{chest CT}, GA-NoEnt predicts \enquote{interval CT scans of the chest}. Although all neural models use pre-trained word embeddings, for Ent and Anonym the multi-word entities do not have pre-trained embeddings since our embeddings are induced on the word level. This may partly explain the competitive performance of NoEnt compared to Ent. We leave the integration of entity embeddings for the future work.

The results for SA reader are far below the performance of GA reader. We also see that it performs much better on anonymized entities than on non-anonymized ones. This is in line with \citet{WangEtAl2017} who find that SA reader suffers a drop of 19 points in exact match on Who-did-what dataset when anonymization is not done. A possible explanation is that anonymization reduces the output space to only several hundred entity candidates for which the output embedding needs to be trained. When we do not use anonymization, the set of output entities increases to the set of all entity types found in all passages, which is several orders of magnitude more. While this effect also occurs for GA reader, it is less pronounced because GA reader scores words in the passage and does not need to learn separate answer word embeddings.

\subsection{Human performance}
To measure the accuracy of human answering, we have used the same sample of data instances as used for the analysis of skills.\footnote{Human answers were collected before the skill analysis.} The queries were answered separately by a novice reader (linguistics background, little-to-none medical knowledge) and by an expert reader (both linguistics and medical background). The annotators needed around 15 minutes on average to read the passage and answer the query. The results are shown at the bottom of Table~\ref{tab:test}. The expert scores higher across all evaluation metrics, with as much as a 7-point advantage in \% F1. This advantage is largely coming from the better performance on those instances where bridging inferences are required (the average F1 score was 10 points higher on these queries), which suggests that domain knowledge is beneficial in the comprehension task. For a novice in a specialized domain, it is harder to build a good situation model that would lead to successful comprehension since it requires more effort---active, strategic processing and establishing ontological relationships in that specific domain. For an expert reader this process is more automatized \citep{KintschAndRawson2008}.

We can see from the table that the best human performance is well below its theoretical upper bound of 100\% F1. An important part of explanation for this lies in the automated dataset construction, which leaves certain queries unanswerable, especially when the authors do not refer to a part in the article but introduce completely new information. Another reason is the problem of \enquote{answer openness}: Typically more than one correct answer is possible and the answers can be correct to various degrees, which we aimed to capture with the use of the embedding metric in the evaluation. Nevertheless, the gap between the best human and machine F1 score is large (around 20 points), leaving considerable space for future applications of machine readers on our dataset.\footnote{For comparison, the gap for SQuAD was 12.2 and for NewsQA 19.8 \citep{TrischlerEtAl2016}.}

\subsection{Breakdown of results by skill}
\begin{figure}
\begin{center}
\includegraphics[width=\columnwidth]{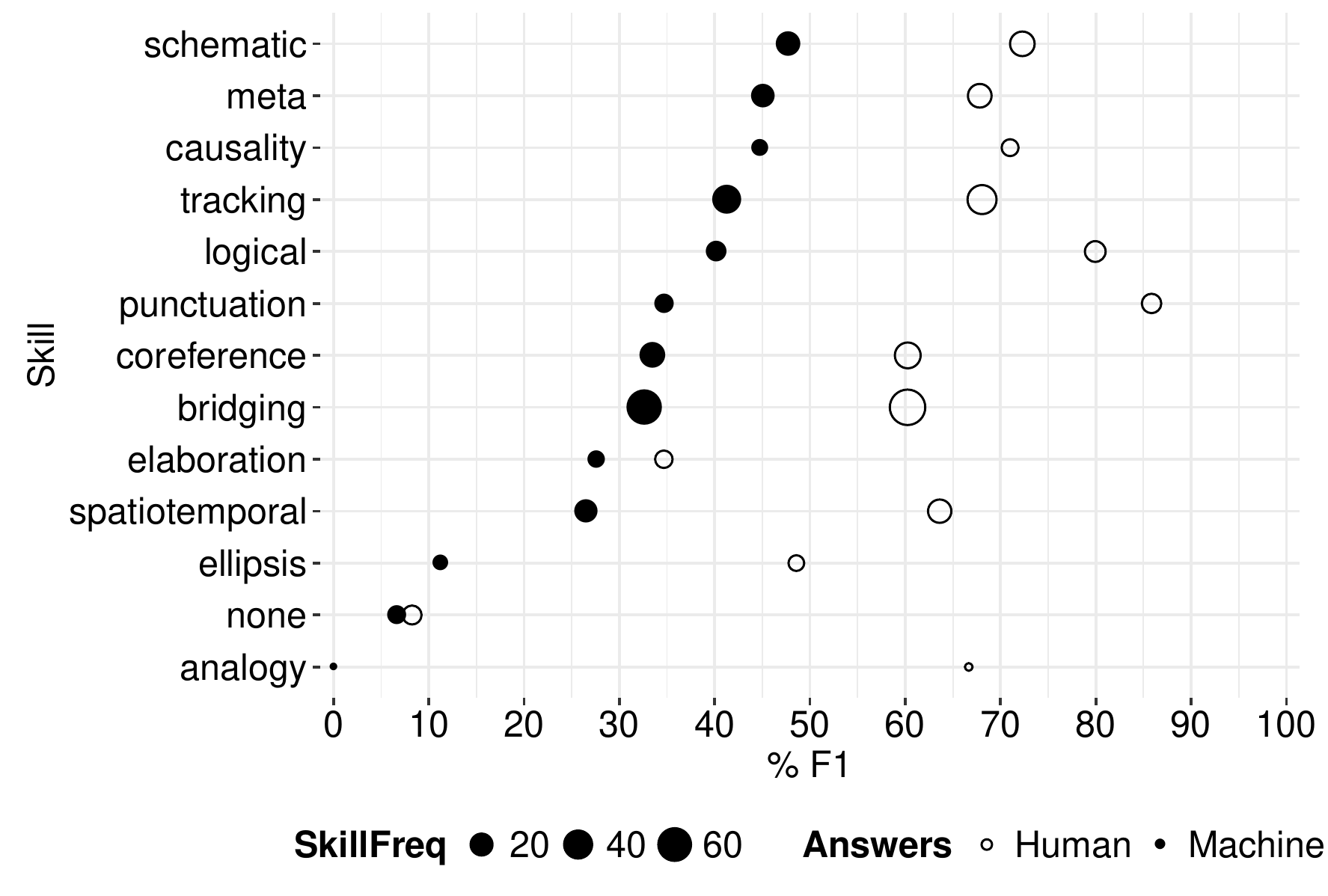}
\caption{Performance per required skill for the human expert and GA-NoEnt reader.}
\label{fig:plotskillsbreakdown}
\end{center}
\end{figure}

To see how the answering performance relates to the skill requirements, we have analyzed the part of the validation set annotated with the skills by averaging F1 values for all instances with a particular skill. In this way, we are able to break down both human and machine performance skill-wise, as shown in Figure~\ref{fig:plotskillsbreakdown}. Because of the small sample size, the results should only be taken as a general indication. The most difficult cases for the GA reader are those annotated with \enquote{none} (unanswerable) and \enquote{ellipsis} (recognizing implicit and omitted information), ignoring \enquote{analogy} for which we only have a single annotated case. Furthermore, spatio-temporal reasoning, elaboration (inferences using general knowledge) and bridging---which is also the most commonly required skill---are the next most difficult ones. The human scores are mostly much higher, which is especially apparent for spatio-temporal reasoning, logical skills and the skill involving punctuation. 
Our findings align with those of \citet{ChuEtAl2017} on the Lambada dataset \citep{PapernoEtAl2016}: Although they used a different categorization of comprehension skills, they also find that GA reader has most difficulties with elaboration (which they refer to as \enquote{external knowledge}), followed by coreference resolution.

\section{Conclusion and future work}
We have introduced a new dataset for domain-specific reading comprehension in which we have constructed around 100,000 cloze queries from clinical case reports. We analyzed the dataset in terms of the skills required for successful comprehension, and applied various baseline methods and state-of-the-art neural readers. We showed that a large gap still exists between the best machine reader and the expert human reader. One direction for future research is improving the reading models on the queries that are currently the most challenging, i.e.\ those requiring world and background domain knowledge. Better representing background knowledge by inducing embeddings for entities or otherwise integrating ontological knowledge is in our opinion a promising avenue for future research.

\section*{Acknowledgments}
We would like to thank Madhumita Sushil and the anonymous reviewers for useful comments. We are also grateful to BMJ Case Reports for allowing the collection of case reports. This work was carried out in the framework of the Accumulate IWT SBO project (nr.\ 150056), funded by the government agency for Innovation by Science and Technology. We also acknowledge the support of the Nvidia GPU Grant Program.

\bibliography{naaclhlt2018}
\bibliographystyle{acl_natbib}

\appendix
\section{Training details and hyper-parameter optimization}\label{app:hyper}

We train the word embeddings using word2vec \citep{MikolovEtAl2013a}, and optimize the window size, the model type (CBOW, skip-gram), the dimensionality and the number of negative samples using random search. For the embedding baseline sim-entity, the evaluation was carried out 20 times on the validation part of our dataset, and we chose the parameter configuration that led to the highest-performing embedding model as measured by F1. We find that higher embedding dimensionality works better, that CBOW obtains somewhat better scores than Skipgram, and that medium-sized word windows work best. The best configuration: 'win\_size': 5, 'min\_freq': 200, 'model': 'cbow', 'dimension': 750, 'neg\_samples': 5. The difference between the lowest and the highest scoring model was 3.4 F1. At prediction time (equation \eqref{eq:simentity}) we set the window size to 3, which worked best on the validation set.


For inclusion in the neural readers, it would be impractical to use the high embedding dimensionality found in the hyper-parameter search from the previous paragraph, so we fix the input embedding dimensionality to 200, as done in \citet{ChenEtAl2016} to keep the training time practical. We optimize the remaining embedding hyper-parameters just like above. The best parameters were: 'win\_size': 4, 'min\_freq': 200, 'model': 'cbow', 'dimension': 200, 'neg\_samples': 9.

For SA reader, we optimized the hidden state size and the dropout rate using 20 different random configurations. The best values were 70 and 0.57, respectively. We explore the same parameters for the GA reader, but add to the search space the feature that indicates the presence of a passage token in the query, which was found useful in the NoEnt set-up.
The best hidden state number and dropout rate were 64 and 0.5, respectively. 
We used the default values for all the remaining hyper-parameters.

\section{List of skills with selected examples}\label{app:skills}
In annotating the skills, we followed the categorization by \citet{SugawaraEtAl2017a}:
\begin{enumerate}
\item Object tracking: tracking or grasping multiple objects; it is a version of list/enumeration skill used in previous skill classifications
\item Mathematical reasoning: whenever a mathematical operation is involved in finding the answer
\item Coreference resolution: direct reference to an object, includes anaphoras. These include inferential processes based on background knowledge or context.
\item Logical reasoning: conditionals, quantifiers, negation, transitivity
\item Analogy: metaphors, metonymy
\item Causal relation: explicit expression such as "why", "the reason of"
\item Spatio-temporal relations
\item Ellipsis: recognizing implicit or omitted information
\item Bridging: inference through grammatical and lexical knowledge (synonymy, idioms etc). 
This link however is not automatic or stereotypical, as in the category of elaboration.
\item Elaboration: inference through commonsense reasoning. 
Note that unlike in the previous category, there is no direct way in which grammatical, lexical or ontological knowledge could help.
\item Meta-knowledge: knowing about the text genre and the main topic being discussed assists in comprehending. In our dataset, knowing the way the queries are constructed (Learning points) is sometimes beneficial.
\item Schematic clause relation: complex sentences that include coordination or subordination
\item Punctuation: understanding parentheses, dashes, quotations, colons etc.
\end{enumerate}

In the following examples, we mark the medical entities in blue, and italicize the parts in the passage that are crucial for answering. Whenever we shorten a part of the passage, we use [...].

\subsection{Bridging inference}
\noindent
\textbf{passage}\\
We report a case of a 72 - year - old Caucasian woman with \textcolor{blue}{pl-7 positive antisynthetase syndrome} . Clinical presentation included \textcolor{blue}{interstitial lung disease} , \textcolor{blue}{myositis} , ‘ mechanic 's hands ’ and \textcolor{blue}{dysphagia} . As \textcolor{blue}{lung injury} was the main concern , \textcolor{blue}{treatment} consisted of \textit{\mbox{\textcolor{blue}{prednisolone}} and \mbox{\textcolor{blue}{cyclophosphamide}}} . Complete remission with reversal of \textcolor{blue}{pulmonary damage} was achieved , as reported by \textcolor{blue}{CT scan} , \textcolor{blue}{pulmonary function tests} and functional status . [...]\\
\textbf{query}\\
Therefore , in severe cases an \textcolor{blue}{aggressive treatment} , combining \textcolor{blue}{\textbf{\underline{\hspace{2cm}}}} and \textcolor{blue}{glucocorticoids} as used in \textcolor{blue}{systemic vasculitis} , is suggested .\\
\textbf{answer}\\
\textcolor{blue}{cyclophoshamide}\\
\textbf{explanation} The reader needs to have the background knowledge that \textcolor{blue}{prednisolone} is a \textcolor{blue}{glucocorticoid}, then it becomes obvious that the answer is \textcolor{blue}{cyclophoshamide}.

\subsection{Object tracking}
\noindent
\textbf{passage}\newline
[...] The patient was managed with \textcolor{blue}{supportive measures} and the National Poisons Information Service was contacted . 
A \textcolor{blue}{toxicology consultant} was involved in view of the unusual mode of administration . 
Although there was no precedent on how to treat a \textcolor{blue}{significant rectal overdose} of \textcolor{blue}{amitriptyline} , \textit{it was advised that the patient be administered a \mbox{\textcolor{blue}{phosphate enema}} and if failed to adequately remove the tablets then the patient should be given \mbox{\textcolor{blue}{whole bowel irrigation}}} with 2 litre of \textcolor{blue}{Klean - Prep} via a \textcolor{blue}{nasogastric tube} . It was also advised that we admit the patient to a high dependency unit and manage him according to the usual protocol for a \textcolor{blue}{tricyclic overdose} if \textcolor{blue}{complications} arose . [...]
\textbf{query}\\
It seems reasonable to attempt careful \textcolor{blue}{removal} of the \textcolor{blue}{drug} from the rectum and if that fails to consider \textcolor{blue}{\textbf{\underline{\hspace{2cm}}}} and \textcolor{blue}{whole bowel irrigation} .\\
\textbf{answer}\\
\textcolor{blue}{phosphate enemas}\\
\textbf{explanation} The query mentions \textcolor{blue}{removal} (A), then \textcolor{blue}{\textbf{\underline{\hspace{2cm}}}} (B) and \textcolor{blue}{whole bowel irrigation} (C). In the passage, one needs to track those elements and choose the right one. This skill should be considered whenever the gap is part of an enumeration or is mentioned as a part of another entity.

\subsection{Meta knowledge}
\noindent
\textbf{query}\\
bedaquiline , a \textcolor{blue}{new agent} with \textcolor{blue}{bactericidal} and sterilising activity against \textcolor{blue}{mycobacterium tuberculosis} , is effective against \textcolor{blue}{\textbf{\underline{\hspace{2cm}}}} when given together with a \textcolor{blue}{background regimen} , and is well tolerated and safe if there is awareness of drug interactions and precautions are taken to avoid \textcolor{blue}{potential qt prolongation} .\\
\textbf{answer}\\
\textcolor{blue}{tuberculosis}\\
\textbf{explanation}
The right answer can be inferred from several parts in the passage (not shown), or even from the title or the query. The query, though, is nowhere in the document explicitly answered.

\end{document}